\PassOptionsToPackage{dvipsnames}{xcolor}
\documentclass[sigconf,nonacm]{acmart}

\usepackage{amsmath,amsfonts}
\usepackage{algorithmic}
\usepackage{algorithm}
\usepackage{array}
\usepackage[caption=false,font=normalsize,labelfont=sf,textfont=sf]{subfig}
\usepackage{textcomp}
\usepackage{stfloats}
\usepackage{url}
\usepackage{verbatim}
\usepackage{graphicx}

\usepackage{xcolor}
\usepackage{booktabs}
\usepackage{multirow}
\usepackage{wrapfig}

\usepackage{makecell}
\usepackage[most]{tcolorbox}
\usepackage{caption}
\usepackage{balance}

\usepackage{colortbl} 

\usepackage{pifont}
\newcommand{\cmark}{\ding{51}}
\newcommand{\xmark}{\ding{55}}
\tcbset{
    enhanced,
    sharp corners, 
    boxrule=0.4pt, 
    colback=white, 
    colframe=black, 
    fonttitle=\bfseries,
    fontupper=\normalsize, 
    left=6pt, right=6pt, top=4pt, bottom=4pt, 
}

\AtBeginDocument{%
  }

\renewcommand\footnotetextcopyrightpermission[1]{}
\setcopyright{none}

\begin{document}

\title{MMCoT${^3}$: Empowering 3D Scene Understanding with 3D-Text Interleaved Multi-Modal Chain-of-Thought Reasoning}

\author{Yongqi Shan}
\affiliation{\institution{Dalian University of Technology}\city{Dalian}\country{China}}
\email{11312327@mail.dlut.edu.cn}
\author{Yifan Wang}
\authornote{Corresponding author.}
\affiliation{\institution{Dalian University of Technology}\city{Dalian}\country{China}}
\email{wyfan@dlut.edu.cn}
\author{Yunzhi Zhuge}
\affiliation{\institution{Dalian University of Technology}\city{Dalian}\country{China}}
\email{zgyz@dlut.edu.cn}
\author{Lijun Wang}
\affiliation{\institution{Dalian University of Technology}\city{Dalian}\country{China}}
\email{ljwang@dlut.edu.cn}
\author{Huchuan Lu}
\affiliation{\institution{Dalian University of Technology}\city{Dalian}\country{China}}
\email{lhchuan@dlut.edu.cn}
\renewcommand{\shortauthors}{Shan et al.}

\begin{abstract}
Understanding and reasoning over 3D scenes from point clouds is essential for a broad range of AI applications. Recent 3D large language models have shown promising results, yet they still face two fundamental limitations: (i) insufficient capability for grounded multi-step reasoning in complex 3D scenes, and (ii) a text-centric inference paradigm that lacks effective interaction between textual reasoning and visual evidence. As a result, these models often fail to verify intermediate inferences, disambiguate similar object instances, and avoid hallucinations. 
To address these limitations, we propose MMCoT${^3}$, the first 3D--Text Interleaved Chain-of-Thought (CoT) framework for 3D scene understanding. The core idea is to explicitly interleave textual reasoning with task-relevant 3D evidence, enabling the model to consult scene observations throughout the reasoning process. In addition, since a textual mention in 3D scenes may correspond to multiple candidate objects, naive object--mention alignment can inject incorrect evidence and mislead subsequent reasoning. 
To mitigate this multi-instance ambiguity, we design an ambiguity-aware evidence aggregation mechanism that uses query-guided cross-attention to fuse candidate object tokens and inject both evidence and ambiguity cues into the reasoning process.
To better equip the model with multi-step reasoning capability, we further develop PointCoT, a large-scale 3D-specific CoT instruction tuning dataset with approximately 176.2K instances spanning captioning, question answering, dialogue, and planning tasks. Extensive experiments on six benchmarks show that our method consistently improves performance and achieves state-of-the-art results, validating the effectiveness of our proposed ambiguity-aware 3D--text interleaved reasoning for 3D scene understanding.
\end{abstract}



\maketitle

\section{Introduction} 

\begin{figure}[t]
    \centering
    \includegraphics[width=\linewidth]{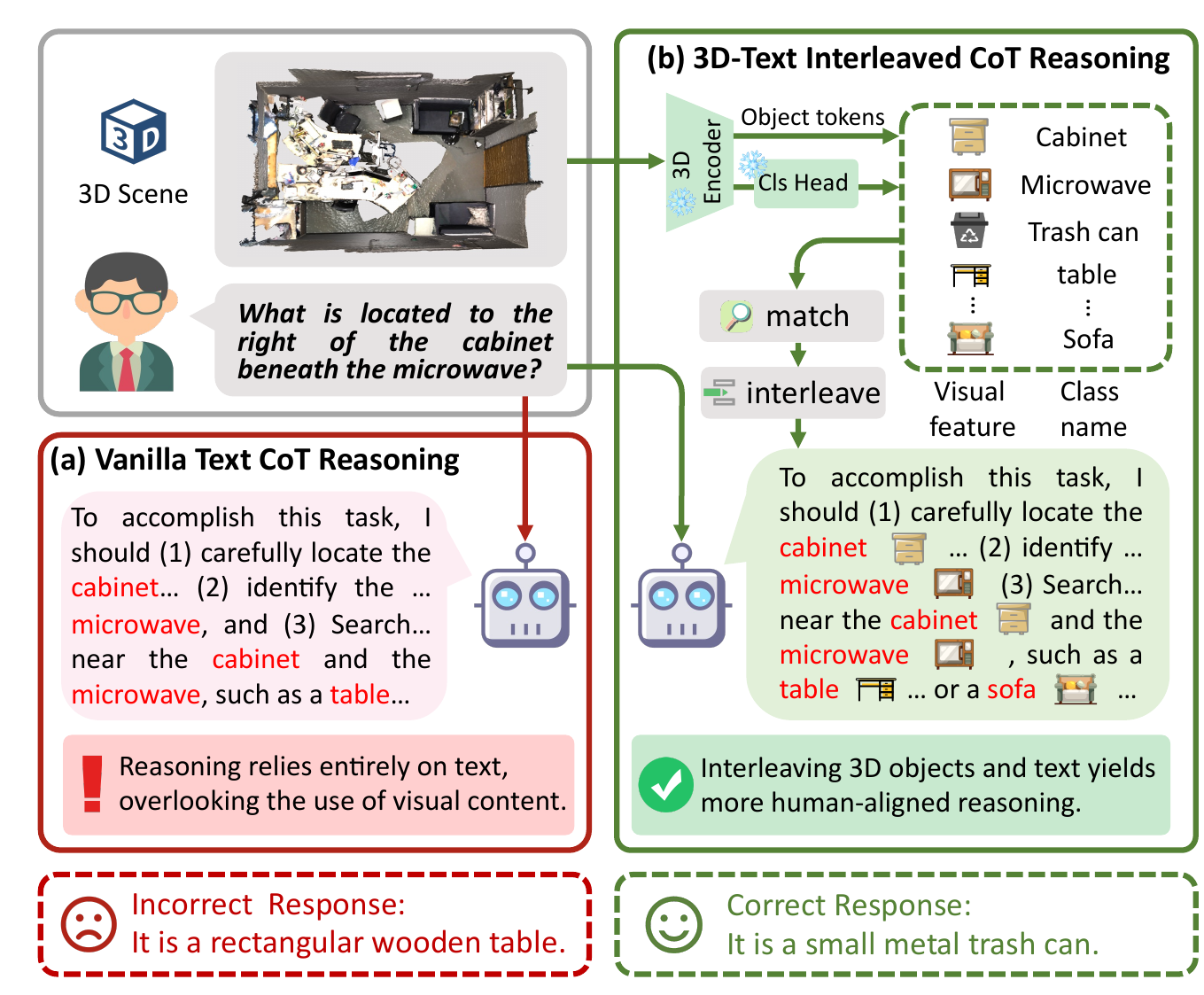} 
    \caption{Comparison between vanilla text-only CoT reasoning and our proposed MMCoT$^3$. (a) Vanilla Text CoT Reasoning is purely text-driven and overlooks visual evidence during reasoning. 
    (b) MMCoT$^3$ introduces 3D--Text interleaved multi-modal CoT reasoning, enabling active consultation and verification of task-relevant 3D evidence during reasoning.}
    \label{fig:teaser}
\end{figure}

Understanding 3D visual information is fundamental to numerous emerging technological domains, including autonomous navigation \cite{arnold2019survey}, augmented and virtual reality \cite{xiong2021augmented}, robotics \cite{huang2023embodied}, and intelligent manufacturing systems \cite{saimon2025advancing}. 
Recent advances  \cite{10981614,hong20233d,xu2024pointllm} have enabled large language models (LLMs) to integrate and reason over rich point-cloud geometry, extending them from text-only modeling to multimodal 3D scene understanding. Representative tasks in 3D scene understanding include 3D captioning~\cite{chen2021scan2cap}, 3D question answering~\cite{azuma2022scanqa,ma2022sqa3d}, 3D dialogue~\cite{huang2023embodied}, and scene-aware task planning~\cite{huang2023embodied}. These tasks provide an important testbed for evaluating whether models can jointly perceive, ground, and reason over complex 3D scenes under natural-language queries.

Existing methods typically employ 3D LLMs \cite{he2024segpoint,wang2023chat} to encode point clouds and text separately, align them in a shared embedding space, and directly generate outputs with the LLM.  
Despite the successes, current methods are still facing two critical issues: 
\textbf{Firstly, insufficient 3D reasoning capability.} The sparse, irregular, and geometry-centric nature of 3D point clouds makes 3D understanding tasks inherently reasoning-intensive. To answer a query, models must jointly reason about object attributes, spatial relations, scene configurations, and task-specific context. However, existing methods still lack the robust reasoning capacity needed for such complex 3D scenes, often leading to incorrect target identification when multiple instances share similar visual characteristics.
\textbf{Secondly, lack of interactions between textual reasoning and visual evidence.} Most existing methods only use 3D representations as a static background context, without actively consulting and verifying them during reasoning (Fig.~\ref{fig:teaser}(a)). Moreover, without effective interaction between textual reasoning and visual evidence, the model cannot properly ground, verify, and revise intermediate conclusions during multi-step inference. Consequently, prior methods are prone to produce hallucinations~\cite{bang2023multitask} in multi-step reasoning. For example, the model may hallucinate non-existent objects or produce spatial reasoning inconsistent with the actual scene layout.

Recently, researchers attempted to reduce the incorrect target identification by improving the selection and reasoning of scene evidence. For example, TSP3D~\cite{guo2025text} introduces a pruning strategy to progressively discard query-irrelevant regions from the 3D scene representation, thereby suppressing redundant scene content and concentrating computation on potentially relevant targets. 
Another research direction is to incorporate Chain-of-Thought (CoT) reasoning to enhance the model's capacity for complex 3D inference, enabling it to better organize and integrate multiple cues, such as object attributes, spatial relations, and global scene context, during multi-step reasoning. 
Nevertheless, existing approaches fall short of fully utilizing visual evidence throughout the reasoning process. As a result, the model cannot effectively revisit 3D evidence to verify intermediate conclusions, refine hypotheses, or resolve ambiguities.

On the other hand, to alleviate the issue of lacking interactions between text and visual information, recent 2D multi-modal reasoning studies~\cite{zhang2025vitcot,gao2025interleaved} have begun to explore multi-modal CoT paradigms that explicitly interleave textual reasoning with visual evidence during inference. Such a paradigm has shown state-of-the-art performance in purely 2D text-centric inference scenarios. Unfortunately, directly transferring this paradigm to 3D scene understanding remains highly nontrivial: (1) there is a lack of large-scale, high-quality 3D-specific CoT dataset, which hinders the further improvements of 3D understanding tasks. 
(2) unlike 2D images/videos, 3D point clouds exhibit a sparse, irregular, and large-scale manner, where only a small fraction of the observed content is relevant to the query. Thus, it is considerably more difficult to identify informative 3D evidence compared with 2D multi-modal reasoning tasks.

A natural idea of alleviating the above limitations is to explicitly interleave intermediate textual reasoning with task-relevant 3D evidence, enabling the model to consult scene observations precisely during multi-step reasoning. 
A naive solution is to align object-level 3D tokens with object mentions in the textual CoT sequence and inject the matched evidence into the corresponding reasoning steps. 
However, such a direct interleaving is inherently brittle in the realistic 3D scenes, where multiple instances frequently share the same category label. 
Thus, a textual entity may match an incorrect token from its associated object tokens, causing subsequent reasoning steps to depend on a mismatched 3D evidence.
Even worse, the ambiguity can further propagate through the CoT process, impairing both the faithfulness of intermediate reasoning and the quality of the final prediction. 
Therefore, effective 3D--Text interleaved reasoning requires not only explicit evidence injection, but also an ambiguity-aware mechanism for identifying relevant candidates, aggregating uncertain evidence, and robustly resolving multi-instance ambiguity.

Following the above analysis, we first develop \textbf{PointCoT}, a large-scale 3D-specific CoT dataset with 176.2K instances, by enriching the LEO instruction-tuning dataset~\cite{huang2023embodied} with high-quality rationales across six representative task settings, including 3D captioning~\cite{chen2021scan2cap}, 3D question answering on ScanNet~\cite{azuma2022scanqa}, 3D question answering on 3RScan~\cite{huang2023embodied}, 3D dialogue~\cite{huang2023embodied}, 3D embodied reasoning~\cite{ma2022sqa3d}, and scene-aware task planning~\cite{huang2023embodied}.
To enhance the capability of responding to various queries, we further propose a difficulty-adaptive CoT generation strategy: for complex queries, we explicitly decompose them into progressive multi-step sub-problems to facilitate structured reasoning. For simple queries, we complement them with concise grounding cues that guide the model toward task-relevant visual evidence.
By providing explicit CoT supervision for diverse 3D tasks, this dataset enables the model to develop stronger multi-step reasoning capability in complex 3D scenes.

To enhance the limited 3D reasoning capability of existing methods and the text-centric inference paradigm, we propose a novel 3D--Text Interleaved Multi-Modal CoT framework \textbf{MMCoT${^3}$} for 3D scene understanding. 
By coupling textual CoT reasoning with 3D evidence, our framework not only strengthens the 3D LLM's capacity for grounded multi-step reasoning, but also moves beyond text-centric inference by enabling the model to actively consult and verify scene observations during intermediate reasoning (Fig.~\ref{fig:teaser}(b)). 

To address the brittleness of direct 3D--Text interleaving in realistic 3D scenes, where multiple instances often share the same semantic category, we further develop an ambiguity-aware evidence aggregation mechanism tailored for 3D scene understanding. 
Instead of prematurely committing to a single object when multiple candidates match a query mention, we aggregate the matched object tokens into a fused evidence representation through query-guided cross-attention. 
Meanwhile, we explicitly estimate the ambiguity of the candidate set and encode it as an additional signal, allowing the model to assess the reliability of the injected evidence during reasoning. 
We then interleave both the task-relevant 3D evidence and its associated ambiguity cue with the textual CoT sequence, enabling the LLM to consult scene observations precisely when intermediate reasoning unfolds, while remaining aware of potential instance-level uncertainty. 
In this way, our framework not only strengthens compositional reasoning over complex 3D scenes, but also establishes an explicit ambiguity-aware evidence-checking process throughout multi-step inference, improving robustness to instance-level ambiguity and mitigating hallucinations.

The contributions of our paper are summarized as follows:

\begin{itemize} 

\item We introduce \textbf{PointCoT}, a large-scale, high-quality 3D-specific CoT dataset built upon LEO instruction~\cite{huang2023embodied} data. To the best of our knowledge, this is the first large-scale 3D-specific CoT dataset for 3D scene understanding, providing explicit supervision for grounded multi-step reasoning across diverse task settings.

\item We propose \textbf{MMCoT${^3}$}, the first 3D-Text Interleaved Multi-Modal CoT framework for 3D
Understanding, which interleaves textual reasoning with task-relevant 3D evidence.

\item We suggest an ambiguity-aware evidence aggregation method to mitigate multi-instance ambiguity, enabling more reliable evidence-grounded reasoning for 3D  understanding.

\item Extensive experiments on six 3D understanding tasks demonstrate that our framework consistently improves performance across captioning, question answering, dialogue, and scene-aware planning, achieving state-of-the-art results (e.g., absolute gains of \textbf{+1.9 CIDEr} on \textbf{ScanQA}, \textbf{+10.9 EM@1} on \textbf{SQA3D}, and \textbf{+2.4 EM@1} on \textbf{3RScanQA}). 

\end{itemize}

\section{Related work}

\begin{figure*}[t]
    \centering
    \includegraphics[width=\linewidth]{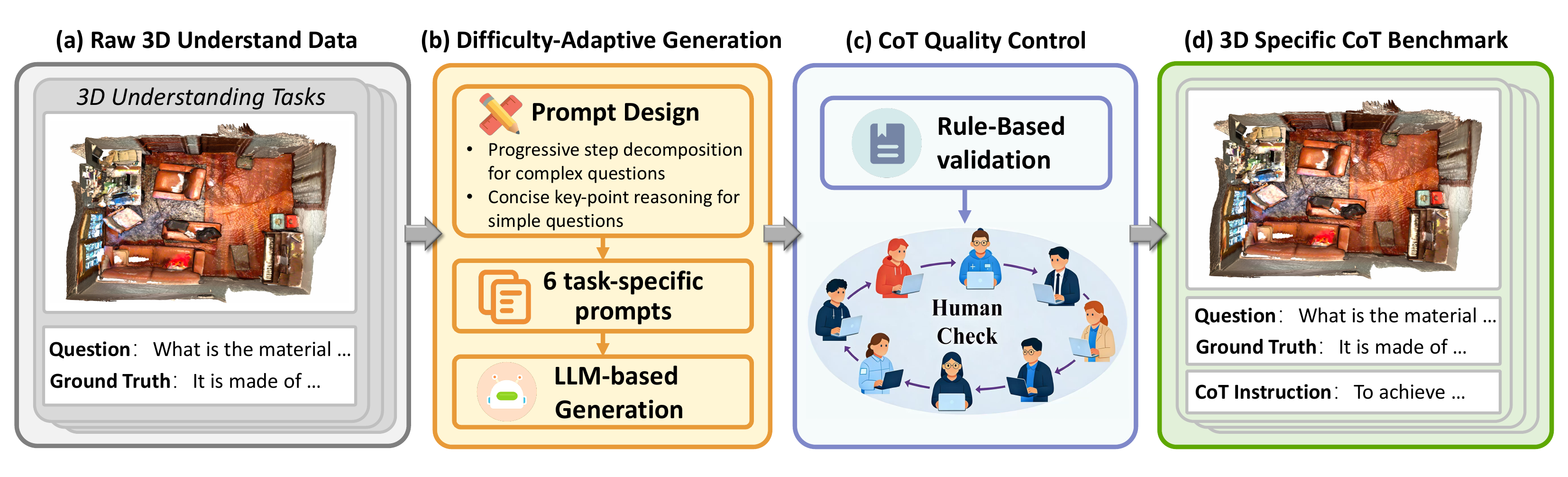} 
    \caption{Construction of the 3D specific CoT benchmark PointCoT: (a) Raw 3D Understanding Data, (b) Difficulty-Adaptive Generation, (c) CoT Quality Control, and (d) the resulting 3D specific CoT benchmark.}
    \label{fig:dataset}
\end{figure*}

\subsection{3D Multi-Modal Large Language Models} 
Recent advances in artificial intelligence have driven rapid progress across visual understanding tasks~\cite{li2026ragtrack,li2026cadtrack,shan2026sceneq,shan2026hdvs}. Advances in multi-modal LLMs have inspired a growing body of work on point-cloud-based 3D large language models~\cite{hong20233d, shi2025multi, li20243dmit, guo2023point, huang2023embodied, yin2025shapegpt, wu2025language, kang2025robin3d, yu2025inst3d, zhang2024spartun3d}. Existing methods mainly focus on bridging 3D representations with language models through feature alignment, unified tokenization, or instruction tuning, as exemplified by 3D-LLM~\cite{hong20233d}, SegPoint~\cite{he2024segpoint}, Point-LLM~\cite{xu2024pointllm}, 3D-LLaVA~\cite{deng20253d}, and 3D-SceneQ~\cite{shan2026sceneq}, which incorporates query-guided adaptive pruning and multi-modal feature enhancement. Other works improve object-centric and embodied understanding by incorporating richer supervision, such as multi-view semantics, inter-object relations, embodied observations, or large-scale instruction data~\cite{yu2025inst3d, huang2023embodied, kang2025robin3d, zhang2024spartun3d}. Meanwhile, OneLLM~\cite{han2024onellm}, Point-Bind LLM~\cite{guo2023point}, and GPT4Point~\cite{qi2024gpt4point} extend this line toward broader multi-modal alignment and generation. Unlike existing 3D LLMs, which primarily focus on representation alignment, instruction tuning, and task adaptation, our work explicitly targets grounded multi-step reasoning in 3D scenes through 3D-specific Chain-of-Thought supervision and 3D--Text interleaved reasoning with task-relevant evidence.

\subsection{Chain-of-Thought Reasoning for Multi-Modal Large Language Models}
Chain-of-Thought (CoT) elicits explicit intermediate reasoning before the final answer and has been shown to improve LLMs’ accuracy and robustness~\cite{wang2022self,wei2022chain,lightman2023let,zhang2024chain}. Following this trend, CoT has been extended to 2D vision-language models~\cite{jaech2024openai,lu2024deepseek,zhang2023multimodal}, where Multi-modal CoT leverages multi-modal rationales to guide prediction and reduce hallucinations~\cite{zhang2023multimodal}, and subsequent Visual CoT works further promote grounded CoT-style visual QA through tailored methods and benchmarks~\cite{chen2024visual,shao2024visual}.
Beyond 2D, CoT-style reasoning has been extended to 3D tasks involving geometry and spatial relations. CoT-Pose~\cite{cha2025cot} maps abstract prompts to detailed pose cues for 3D pose generation, while CoT3DRef~\cite{abdelrahman2023cot3dref} predicts an anchor chain before the final target for data-efficient 3D grounding. AffordBot~\cite{wang2025affordbot} further uses multi-modal CoT to select informative views and reason step-by-step about 3D affordances and interactions, and MM-Spatial~\cite{daxberger2503mm} leverages metric depth via CoT use at inference. 
Despite recent progress, existing CoT-enabled 3D LLMs largely remain within text-only reasoning chains, treating 3D observations mainly as static prefix context, while failing to fully exploit visual evidence for intermediate reasoning. In contrast, we argue that 3D–text interleaved multi-modal CoT is crucial for faithful spatial reasoning, as it can explicitly couple intermediate deductions with geometric cues and multi-view context. Additionally, to support systematic evaluation, we build a 3D CoT benchmark for assessing the effectiveness of 3D–text interleaved reasoning.

\section{3D-Specific Chain-of-Thought Dataset}
To alleviate the scarcity of Chain-of-Thought (CoT) supervision in current 3D scene understanding research, we develop a novel 3D-specific CoT benchmark, termed \textbf{PointCoT}. 

As illustrated in Fig.~\ref{fig:dataset}(a), PointCoT is built by augmenting existing 3D instruction data~\cite{huang2023embodied} with high-quality Chain-of-Thought rationales, aiming to support grounded multi-step reasoning over point-cloud scenes. 
As summarized in Table~\ref{tab:dataset_stats_scene}, PointCoT contains 176.2K samples spanning six task settings across different scenes, tasks, and difficulty levels, including 3D captioning~\cite{chen2021scan2cap}, 3D question answering on ScanNet~\cite{azuma2022scanqa}, 3D question answering on 3RScan~\cite{huang2023embodied}, 3D dialogue~\cite{huang2023embodied}, 3D embodied reasoning~\cite{ma2022sqa3d}, and scene-aware task planning~\cite{huang2023embodied}.
The underlying 3D scenes are drawn from two representative indoor datasets, \textit{ScanNet} and \textit{3RScan}, covering a total of 2,995 indoor scenes. 
As summarized in Table~\ref{tab:dataset_stats_scene}, PointCoT contains 176.2K samples spanning six task settings with detailed statistics across scenes, tasks, and difficulty levels. 
For each sample, PointCoT provides a 3D scene, a language instruction or query, a CoT rationale, and the corresponding target response, thereby enabling models to learn how to align intermediate reasoning steps with task-relevant 3D evidence.

\begin{table*}[t]
\centering
\caption{Statistics of the proposed 3D-specific CoT dataset PointCoT across datasets and tasks. Compared with prior 3D instruction resources, PointCoT provides explicit difficulty-aware CoT supervision for grounded multi-step reasoning.}
\vspace{-5pt}
\label{tab:dataset_stats_scene}
\renewcommand{\arraystretch}{1.1}
\setlength{\tabcolsep}{3pt}
\small
\begin{tabular}{l c c c c c c c c}
\toprule
\multirow{2}{*}{\textbf{Dataset}}
& \multirow{2}{*}{\shortstack[c]{\textbf{3D Captioning~\cite{chen2021scan2cap}}}}
& \multirow{2}{*}{\shortstack[c]{\textbf{3D QA~\cite{azuma2022scanqa}}}}
& \textbf{3D Embodied}
& \multirow{2}{*}{\shortstack[c]{\textbf{3D Dialogue~\cite{huang2023embodied}}}}
& \textbf{Task~\cite{huang2023embodied}}
& \textbf{Difficulty}
& \textbf{Chain-of-Thought}
& \multirow{2}{*}{\textbf{Samples}} \\
& & & \textbf{Reasoning~\cite{ma2022sqa3d}} & & \textbf{Planning} & \textbf{Aware} & \textbf{Supervision} & \\
\midrule
\textbf{ScanNet~\cite{dai2017scannet}} 
& \cmark & \cmark & \cmark & \xmark & \xmark & \xmark & \xmark & 88.9K \\

\textbf{3RScan~\cite{wald2019rio}} 
& \xmark & \cmark & \xmark & \cmark & \cmark & \xmark & \xmark & 87.3K \\

\textbf{Leo-instruct~\cite{huang2023embodied}} 
& \cmark & \cmark & \cmark & \cmark & \cmark & \xmark & \xmark & 176.2K \\
\midrule
\textbf{PointCoT (Ours)} 
& \cmark & \cmark & \cmark & \cmark & \cmark & \cmark & \cmark & 176.2K \\
\bottomrule
\end{tabular}
\end{table*}

\subsection{Difficulty-Adaptive Generation}

As illustrated in Fig.~\ref{fig:dataset}(b), to support CoT data construction across diverse 3D understanding tasks, we develop a difficulty-adaptive data generation strategy, rather than relying on a single unified prompting scheme. 
Specifically, we propose six prompt templates according to the characteristics of different datasets and task types, and use Deepseek-R1~\cite{guo2025deepseek} to generate CoT supervision for each sample. The key principle is to align the prompt design with the reasoning demands of the target task: for complex tasks, the prompts explicitly encourage progressive multi-step decomposition, whereas for simple tasks, they instead emphasize concise grounding cues. 

For complex tasks, such as dialogue and scene-aware planning, the prompts encourage the model to decompose the problem into logical sub-steps, identify relevant context, and progressively organize the response. This leads to more structured supervision for grounded multi-step reasoning. 
In contrast, for simpler tasks such as question answering and captioning, we design prompts that emphasize compact grounding cues, including target objects, reference objects, spatial relations, and discriminative attributes. Such prompts encourage the model to produce compact rationales that remain closely tied to the 3D scene. Table~\ref{tab:dataset_stats_scene} further shows that PointCoT covers both easy and complex settings, reflecting the diverse reasoning demands of different 3D understanding tasks.

\subsection{Chain-of-Thought Quality Control}

As illustrated in Fig.~\ref{fig:dataset}(c)--(d), we further perform a post-hoc quality control procedure to ensure the reliability of the generated CoT annotations. This procedure includes two stages: rule-based validation and human verification.
In the first stage, we perform rule-based validation to identify samples with common failure patterns, including: (1) structural defects, such as missing reasoning content, severe redundancy, or format inconsistency; and 
(2) semantic mismatch, where the generated rationale is weakly related to the given instruction or query. 
In the second stage, we conduct human verification by asking six volunteers to each review 3,000 randomly sampled instances from each sub-dataset. The verification is based on three criteria: (1) query relevance, namely whether the generated rationale is well aligned with the given instruction or query; (2) reasoning coherence, namely whether the rationale is logically consistent and follows a plausible reasoning flow; and (3) scene groundedness, namely whether the rationale is reasonably supported by the target 3D scene, including relevant objects, attributes, spatial relations, and contextual cues. 

\begin{figure}[t]
    \centering
    \scalebox{1}{
    \includegraphics[width=\linewidth]{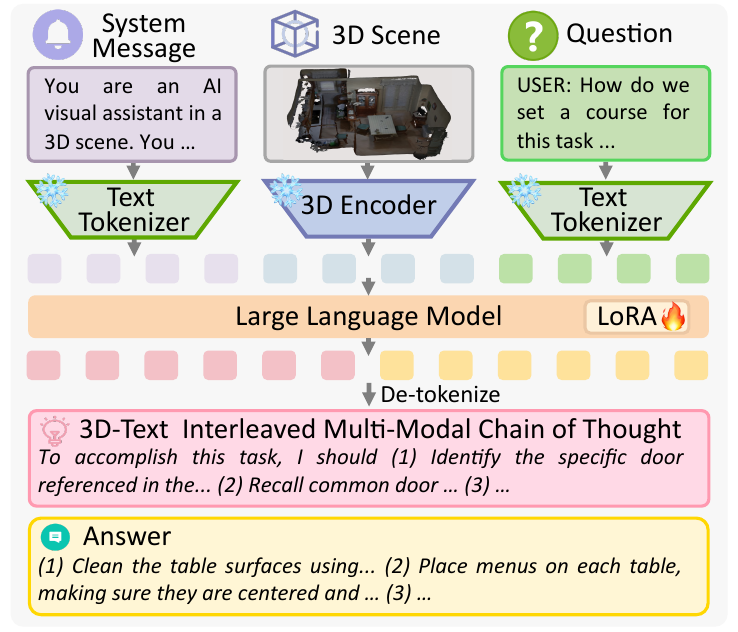} }
    \caption{Overview of the MMCoT$^3$ framework, which integrates a 3D point-cloud scene, the user's textual question, and a system message to generate 3D--Text interleaved multi-modal chain-of-thought and informed responses.}
    \label{fig:embedding}
\end{figure}

\section{Method}
To address the limitations of previous 3D understanding methods, as shown in Fig.~\ref{fig:embedding}, we introduce a novel 3D understanding paradigm called MMCoT${^3}$. Specifically, MMCoT${^3}$ primarily consists of four steps: candidate selection (Sec.~\ref{sec:candidate}), query-based token fusion (Sec.~\ref{sec:fusion}), ambiguity modeling (Sec.~\ref{sec:ambiguity}), and ambiguity-aware evidence injection (Sec.~\ref{sec:injection}).

\subsection{Overview}\label{sec:overview}

Given a point-cloud 3D scene and a question $q$, MMCoT$^3$ aims to generate a grounded response consisting of a CoT rationale and the final answer. MMCoT$^3$ is trained in two stages: (1) \emph{3D vision-language alignment}, which maps object-centric 3D representations into the LLM feature space, and (2) \emph{instruction tuning}, which equips the model with 3D-text interleaved reasoning ability through ambiguity-aware CoT supervision.

In the 3D vision-language alignment stage, we encode the input scene with an object-centric 3D encoder, PointNet++~\cite{qi2017pointnet++}. For each object $o_i$, the encoder extracts a geometric embedding, which is further processed by a Spatial Transformer~\cite{chen2022language} to model inter-object interactions and produce object-centric 3D tokens. A linear projection then maps these tokens into the LLM feature space, yielding the final 3D object representation $f_{3D} \in \mathbb{R}^{N \times d}$, where $N$ is the number of objects and $d$ is the LLM hidden dimension. Using the system prompt, question $q$, and $f_{3D}$ as input, the model learns a shared multi-modal feature space for grounded generation. Meanwhile, textual tokens are projected into the same feature space through a trained text projection layer, so that 3D object representations and textual representations are aligned for downstream reasoning.

In the instruction tuning stage, rather than treating 3D features as static prefix context, MMCoT$^3$ injects task-relevant 3D evidence into intermediate reasoning. Specifically, given a textual CoT, we identify object-related mentions, retrieve candidate 3D object tokens from the scene (Sec.~\ref{sec:candidate}), fuse ambiguous candidates via query-based token fusion (Sec.~\ref{sec:fusion}), estimate the ambiguity of the fused evidence (Sec.~\ref{sec:ambiguity}), and inject the resulting evidence and ambiguity tokens into the corresponding reasoning positions (Sec.~\ref{sec:injection}). This yields an ambiguity-aware 3D-text interleaved multi-modal CoT, enabling the model to condition intermediate reasoning on object-level 3D evidence. 

For the $b$-th training sample, we denote by $\tilde{C}^{(b)}$ the ambiguity-aware interleaved multi-modal CoT constructed by injecting the fused evidence and ambiguity tokens into the corresponding reasoning positions (detailed in Sec.~\ref{sec:injection}). The final augmented response sequence is then defined as:
\begin{equation}
\hat{s}_{\mathrm{res}}^{(b)} = \tilde{C}^{(b)} \oplus A^{(b)},
\end{equation}
where $A^{(b)}$ is the ground-truth answer.
The training objective is:
\begin{equation}
\mathcal{L}(\theta, \mathcal{B})
=
- \sum_{b=1}^{|\mathcal{B}|} \sum_{t=1}^{\hat{T}_b}
m^{(b,t)}
\log p_{\theta}\!\left(
\hat{s}_{\mathrm{res}}^{(b,t)}
\mid
\hat{s}_{\mathrm{res}}^{(b,<t)},\,
s_{\mathrm{prefix}}^{(b)}
\right),
\end{equation}
where $s_{\mathrm{prefix}}^{(b)}$ consists of the system prompt, the question, and the input 3D object representations, and $\hat{T}_b$ denotes the length of the augmented response sequence $\hat{s}_{\mathrm{res}}^{(b)}$. Here, $m^{(b,t)}=1$ for textual target tokens, including both the textual CoT and the answer, and $m^{(b,t)}=0$ for injected visual tokens. In this way, the injected multi-modal evidence participates in the forward self-attention computation, while the loss is imposed only on textual outputs.

\begin{figure*}[t]
    \centering
    \scalebox{1}{
    \includegraphics[width=\linewidth]{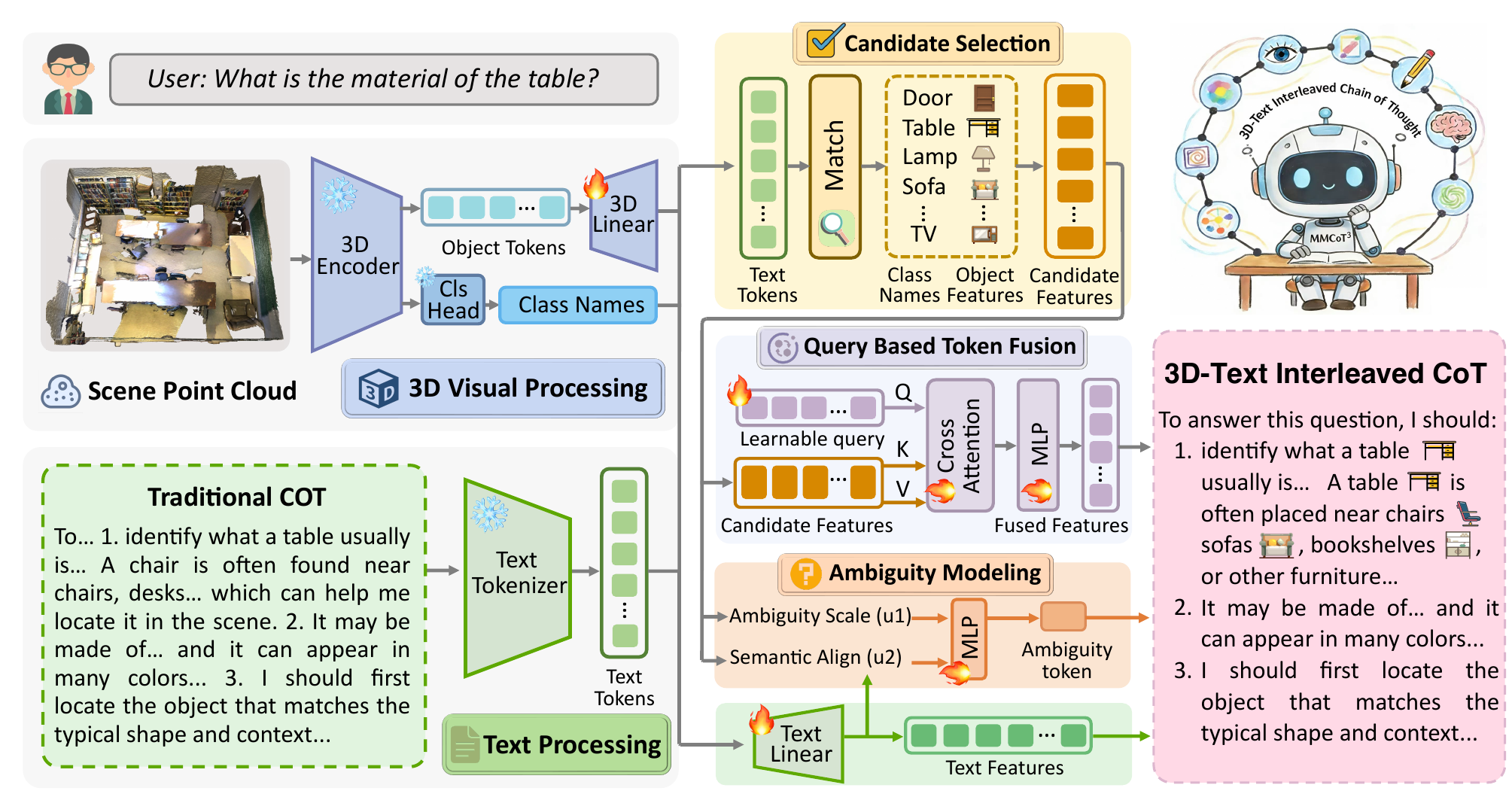}}
    \caption{Construction of the 3D-text interleaved multi-modal CoT. Starting from a scene point cloud and a textual CoT, the framework first matches object-related mentions with candidate 3D objects, then performs query-based token fusion and ambiguity modeling to obtain evidence and ambiguity cues. These tokens are injected into the reasoning chain to form the final 3D-text interleaved multi-modal CoT.}
    \label{fig:pipeline}
\end{figure*}

\subsection{Candidate Selection}\label{sec:candidate}

As shown in Fig.~\ref{fig:pipeline}, during instruction tuning, we use the frozen PointNet++~\cite{qi2017pointnet++} encoder, together with its classification head and the trained 3D projection layer from the alignment stage, to obtain object-level 3D candidates from the input scene.

Given a 3D point-cloud scene, the encoder extracts object-centric geometric tokens, while the classification head predicts a category name $y_i$ for each object. The object features are then projected into the aligned LLM feature space, yielding $\{o_i\}_{i=1}^{N}$, where $N$ is the number of objects in the scene and $o_i \in \mathbb{R}^{d}$ denotes the aligned representation of the $i$-th object.

To associate textual reasoning steps with scene objects, we perform lexical candidate matching in the tokenized text space. Specifically, we tokenize each predicted category name $y_i$ as $\tau(y_i)$ and the CoT as $\tau(C) = [c_1, c_2, \dots, c_T]$. Any span in $\tau(C)$ whose token sequence exactly matches $\tau(y_i)$ is treated as an object-related mention $m$. The candidate set for $m$ is then defined as:
\begin{equation}
\mathcal{O}_m = \{\, o_i \mid \tau(y_i) = \tau(m) \,\},
\end{equation}
where $\tau(m)$ denotes the token sequence of the matched mention, and $K = |\mathcal{O}_m|$ is the number of matched candidates.

For each mention, this step yields a candidate set of object tokens whose predicted category names match the tokenized form of the corresponding CoT span. The resulting candidates are then used for subsequent query-based fusion and ambiguity modeling.

\subsection{Query-Based Token Fusion}\label{sec:fusion}

Given the candidate set $\mathcal{O}_m=\{o_i\}_{i=1}^{K}$ for a textual CoT mention $m$, we construct a visual evidence token $f_m$ by query-based token fusion.
If $K=1$, the mention matches a single object token, and we directly use it as the visual evidence $f_m = o_1$.

If $K>1$, multiple object tokens share the same predicted category name. 
We therefore use a global learnable query vector $q \in \mathbb{R}^{d}$, shared across mentions, to aggregate the candidate set through cross-attention, yielding a compact fused evidence token:
\begin{equation}
f_m = \mathrm{MLP}\big(\mathrm{CrossAttn}(q, O_m)\big),
\end{equation}
where $O_m = [o_1,\dots,o_K] \in \mathbb{R}^{K \times d}$ is the candidate object set for mention $m$. 
The resulting token $f_m \in \mathbb{R}^{d}$ is then passed to the subsequent ambiguity modeling module for reliability estimation.

\subsection{Ambiguity Modeling}\label{sec:ambiguity}

Given the fused evidence token $f_m$ for a textual CoT mention $m$, we further estimate its ambiguity before injecting it into the reasoning chain. 
We model ambiguity from two aspects: candidate-set ambiguity and semantic inconsistency.
To quantify ambiguity induced by multiple same-category candidates, we define a candidate-set ambiguity term based on the set size:
\begin{equation}
u_1 = \log(1 + K),
\end{equation}
where $K = |\mathcal{O}_m|$ is the number of object tokens matched to $m$.

The tokenized CoT sequence $\tau(C)$ is processed by the trained text projection layer from the 3D vision-language alignment stage, yielding aligned textual features
$H = [h_1, h_2, \dots, h_T]$, where $h_j \in \mathbb{R}^{d}$.
For a matched mention $m$ spanning positions $[l_m, r_m]$, we obtain its textual representation $t_m$ by mean-pooling the corresponding aligned textual features:
\begin{equation}
t_m = \frac{1}{r_m-l_m+1} \sum_{j=l_m}^{r_m} h_j.
\end{equation}
Since both $f_m$ and $t_m$ lie in the aligned LLM feature space, we define the semantic inconsistency term as:
\begin{equation}
u_2 = 1 - \cos(f_m, t_m),
\end{equation}
where a larger $u_2$ indicates lower semantic consistency between the fused visual evidence and the corresponding textual mention.

Finally, we combine the two terms to obtain the ambiguity token:
\begin{equation}
a_m = \mathrm{MLP}_{\mathrm{amb}}([u_1, u_2]),
\end{equation}
where $[\cdot,\cdot]$ denotes concatenation and $a_m \in \mathbb{R}^{d}$ denotes the ambiguity token in the aligned LLM feature space.

\subsection{Ambiguity-Aware Evidence Injection}\label{sec:injection}

We insert the fused evidence token $f_m$ and ambiguity token $a_m$ into the aligned textual feature sequence $H$. For a mention $m$ spanning $[l_m, r_m]$, the augmented sequence is:
\begin{equation}
\tilde{H}_m =
[h_1,\dots,h_{l_m-1},\, f_m,\, a_m,\, h_{l_m},\dots,h_{r_m},\dots,h_T].
\end{equation}

For multiple matched mentions, we apply the insertion in left-to-right order to obtain the final interleaved feature sequence $\tilde{H}$. This sequence corresponds to the ambiguity-aware interleaved multi-modal CoT $\tilde{C}$ used to construct the augmented response sequence in the overview, and is fed into the LLM for autoregressive reasoning and response generation.

\tcbset{
    enhanced,
    sharp corners, 
    boxrule=0.4pt, 
    colback=white, 
    colframe=black, 
    fonttitle=\bfseries,
    fontupper=\normalsize, 
    left=6pt, right=6pt, top=4pt, bottom=4pt, 
}

\begin{figure*}[t]
    \centering
    \scalebox{0.85}{
    \includegraphics[width=\linewidth]{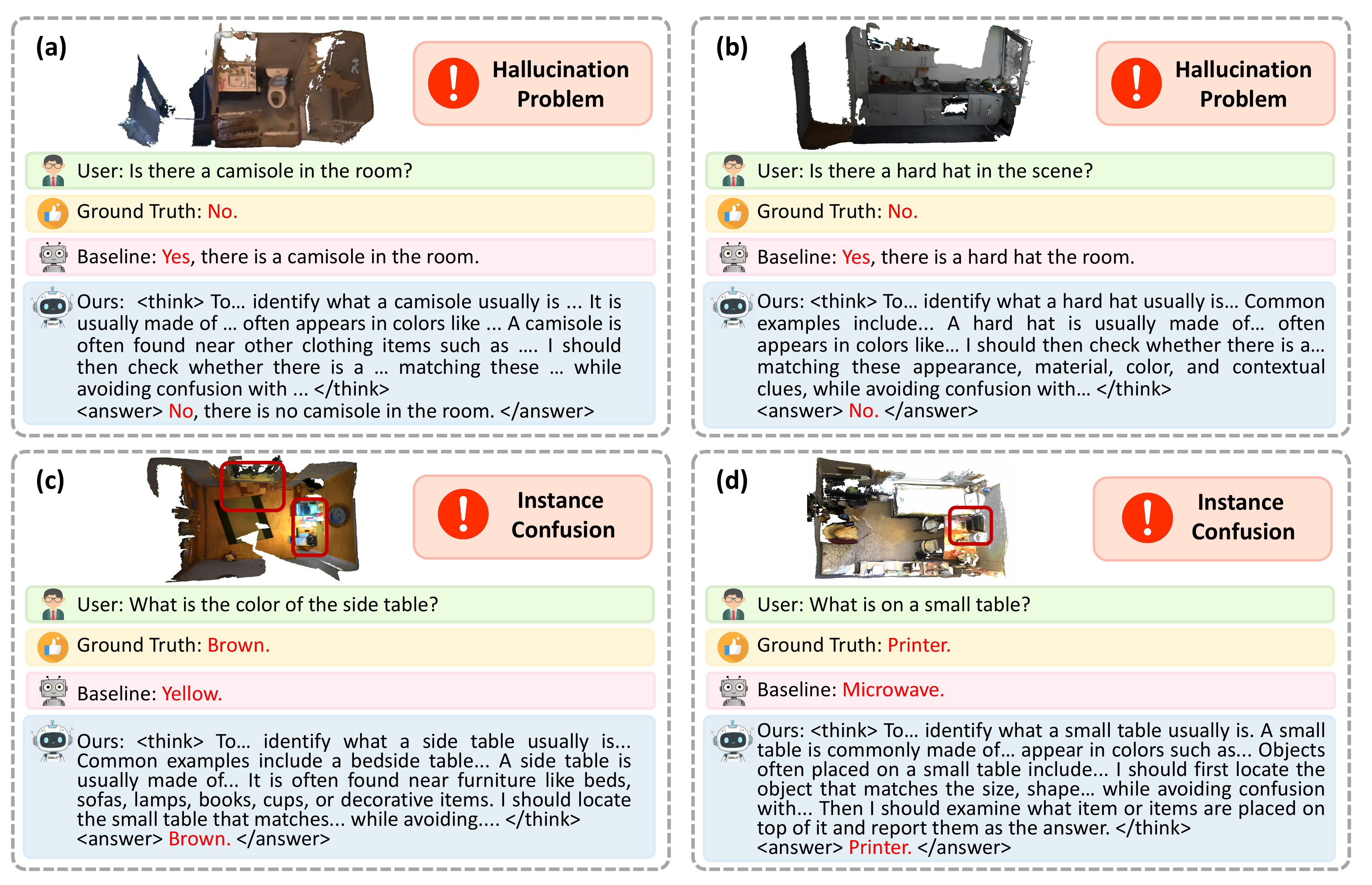}}
    \vspace{-2mm}
    \caption{Qualitative comparison between the baseline and MMCoT$^3$. (a)–(b) show that MMCoT$^3$ suppresses hallucinations of nonexistent objects, while (c)–(d) highlight its improved ability to disambiguate visually similar instances.}
    \label{fig:view}
\end{figure*}

\section{Experiments}

\subsection{Experiment Setup}

\textbf{Datasets.}
MMCoT$^3$ is trained in two stages and evaluated on four representative 3D scene understanding tasks: captioning~\cite{chen2021scan2cap}, question answering~\cite{azuma2022scanqa, ma2022sqa3d}, dialogue~\cite{huang2023embodied}, and scene-aware planning~\cite{huang2023embodied}. 
For 3D vision-language alignment, we initialize the model from the released first-stage checkpoint of LEO~\cite{huang2023embodied}, which is trained on object-, context-, and scene-level 3D language supervision. 
For instruction tuning, we use \textbf{PointCoT}, our proposed 3D-specific CoT dataset with approximately 176.2K instances. Each sample comprises a 3D scene, a task-specific language input, a chain-of-thought rationale, and the target response, where the language input varies across tasks and may include a question, a situation description, or dialogue context. 
We evaluate the model under both descriptive and reasoning-intensive settings.

\textbf{Evaluation Metrics.}
We use standard metrics to evaluate different task types. 
For language generation tasks, we report CIDEr, BLEU, METEOR, and ROUGE. 
For open-ended generation tasks where multiple valid outputs may exist, we additionally report sentence similarity~\cite{reimers2019sentence} to better measure semantic alignment. 
For question answering tasks requiring exact factual correctness, we report top-1 exact match accuracy (EM@1).

\textbf{Implementation Details.}
We adopt Vicuna-7B~\cite{chiang2023vicuna} as the language backbone and initialize the model from the released first-stage checkpoint of LEO~\cite{huang2023embodied}. 
We keep the PointNet++~\cite{qi2017pointnet++} 3D encoder and the language tokenizer frozen during training, while fine-tuning the language backbone with LoRA~\cite{hu2021lora} using a low-rank setting of $r=16$. 
The trainable parameters include the spatial transformer, the 3D projection layer, the text projection layer, the query-based fusion module, and the ambiguity modeling module. 
We train the model for 10 epochs using the AdamW optimizer with a learning rate of $3\times10^{-5}$. All experiments are conducted on two NVIDIA A800 GPUs.

\subsection{Comparison with State-of-the-Art Methods}

We evaluate MMCoT$^3$ on six 3D scene understanding benchmarks built on ScanNet and 3RScan. As shown in Tables~\ref{tab:scannet} and~\ref{tab:rscan}, MMCoT$^3$ achieves consistently strong performance across captioning, question answering, situated reasoning, dialogue, and planning, demonstrating robust generalization across both tasks and datasets.

As shown in Table~\ref{tab:scannet}, on the ScanNet-based~\cite{dai2017scannet} benchmarks, MMCoT$^3$ sets a new state of the art on all five Scan2Cap~\cite{chen2021scan2cap} metrics, achieving 79.9 CIDEr, 44.3 BLEU-4, 31.2 METEOR, 65.3 ROUGE, and 63.8 Sim. Compared with the strong 3D LLM baseline LEO, this corresponds to gains of 7.5, 6.1, 3.3, 7.2, and 8.5 points, respectively. On ScanQA, MMCoT$^3$ achieves the best CIDEr and METEOR scores, while remaining competitive on BLEU-4 and ROUGE. This pattern suggests that MMCoT$^3$ mainly improves semantic grounding and reasoning quality, whereas lexical overlap metrics are less directly aligned with the gains brought by evidence-grounded multi-step reasoning. On SQA3D, MMCoT$^3$ reaches 65.4 EM@1, exceeding the best prior result by 10.9 points, further verifying its advantage on more challenging situated reasoning.

The gains further generalize to the 3RScan-based~\cite{wald2019rio} benchmarks. As shown in Table~\ref{tab:rscan}, MMCoT$^3$ consistently surpasses both the instruction-tuned baseline and LEO across planning, dialogue, and question answering, achieving 83.9 Sim on 3RScanPlanning, 84.1 Sim on 3RScanDialog, and 54.3 EM@1 on 3RScanQA. Compared with LEO, the improvements are 2.8, 10.8, and 2.4 points, respectively. The especially pronounced gain on 3RScanDialog highlights the advantage of our framework for more compositional and context-dependent reasoning, where accurate grounding of intermediate reasoning steps is particularly important.

Overall, these results suggest that the gains of MMCoT$^3$ arise from a stronger evidence-grounded reasoning paradigm rather than task-specific engineering. For complex tasks such as planning and dialogue, 3D--Text interleaving enables the model to decompose long-horizon decisions into grounded intermediate steps and continuously verify them against the scene, thereby reducing error accumulation. For relatively simpler tasks such as captioning and question answering, it still provides consistent benefits by strengthening object-level grounding and improving the recognition of attributes and spatial relations, leading to more accurate and semantically faithful predictions.

\renewcommand{\arraystretch}{1.0}
\begin{table*}[t]
\centering
\caption{Comparison of MMCoT$^3$ with state-of-the-art models on Scan2Cap, ScanQA, and SQA3D benchmarks across various evaluation metrics. ``EM@1" denotes top-1 exact match accuracy. Scan2Cap n-gram metrics use IoU $\ge$ 0.5. \textsuperscript{\textdagger} is GPT-3 answers based on generated captions. “Sim” denotes sentence similarity.}
\label{tab:scannet}
\vspace{-2mm}
\resizebox{1\linewidth}{!}{
\begin{tabular}{l>{\centering\arraybackslash}p{1cm}>{\centering\arraybackslash}p{1cm}>{\centering\arraybackslash}p{1cm}>{\centering\arraybackslash}p{1cm}>{\centering\arraybackslash}p{1cm}>{\centering\arraybackslash}p{1cm}>{\centering\arraybackslash}p{1cm}>{\centering\arraybackslash}p{1cm}>{\centering\arraybackslash}p{1cm}>{\centering\arraybackslash}p{1.5cm}}
\toprule
\multirow{3.5}{*}{Method} & \multicolumn{5}{c}{\raisebox{-1.5ex}{Scan2Cap (val)}} & \multicolumn{4}{c}{\raisebox{-1.5ex}{ScanQA (test)}} & SQA3D (test) \\
\cmidrule(lr){2-6} \cmidrule(lr){7-10} \cmidrule(lr){11-11}
& CIDEr & BLEU-4 & METEOR & ROUGE & Sim & CIDEr & BLEU-4 & METEOR & ROUGE & EM@1 \\
\midrule
\textit{Task-specific models} \\
Scan2Cap~\cite{chen2021scan2cap} & 35.2 & 22.4 & 21.4 & 43.5 & - & - & - & - & - & 41.0\textsuperscript{\textdagger} \\
Vote2Cap-DETR~\cite{chen2023end} & 61.8 & 34.5 & 26.2 & 54.4 & - & - & - & - & - & - \\
ScanRefer+MCAN~\cite{yu2019deep} & - & - & - & - & - & 55.4 & 7.9 & 11.5 & 30.0 & - \\
ScanQA~\cite{azuma2022scanqa} & - & - & - & - & - & 64.9 & 10.1 & 13.1 & 33.3 & 47.2 \\
\midrule
\textit{Task-specific fine-tuned} \\
Chat-3D~\cite{wang2023chat}& - & - & - & - & - & 53.2 & 6.4 & 11.9 & 28.5 & - \\
3D-VisTA~\cite{zhu20233d}& 66.9 & 34.0 & 27.1 & 54.3 & 53.8 & 69.6 & 10.4 & 13.9 & 35.7 & 48.5 \\
3D-LLM (FlanT5)~\cite{hong20233d} & - & - & - & - & - & 69.4 & 12.0 & 14.5 & 35.7 & - \\
NaviLLM~\cite{zheng2024towards}& - & - & - & - & - & 75.9 & 12.5 & 15.4 & 38.4 & - \\
Inst3D-LMM~\cite{yu2025inst3d} & 79.7& 38.3& 27.5& 57.2 & - & 88.6 & 14.9 & 18.4 & 42.6 & - \\
3UR-LLM~\cite{10981614} & -& -& -& - & - & 87.7 & 15.5 & 18.4 & 41.5 & - \\
3D-LLaVA~\cite{deng20253d} & 78.8& 36.9& 27.1& 57.7 & - & 92.6 & \textbf{17.1} & 18.4 & 43.1 & 54.5 \\
LEO~\cite{huang2023embodied} & 72.4 & 38.2 & 27.9 & 58.1 & 55.3 & 101.4 & 13.2 & 20.0 &  \textbf{49.2} & 50.0\\
\midrule
MMCOT$^3$ & \textbf{79.9} &  \textbf{44.3} &  \textbf{31.2} &  \textbf{65.3} &  \textbf{63.8} &  \textbf{103.3} &  15.3 &  \textbf{20.6} & 48.9 &  \textbf{65.4} \\
\bottomrule
\end{tabular}}
\end{table*}

\renewcommand{\arraystretch}{1}
\begin{table}[t]
\centering
\caption{Comparison of MMCoT$^3$ with baselines on 3RScan benchmarks. We report exact match accuracy (EM@1) for QA tasks and sentence similarity (Sim) for non-QA tasks.}
\vspace{-2mm}
\label{tab:rscan}
\resizebox{1\linewidth}{!}{
\small
\begin{tabular}{lccc}
\toprule
\multirow{2}{*}{Method} 
& 3RscanPlanning 
& 3RscanDialog
& 3RscanQA \\
\cmidrule(lr){2-2} \cmidrule(lr){3-3} \cmidrule(lr){4-4}
& Sim & Sim & EM@1 \\
\midrule
Baseline & 53.9& 61.4& 39.6\\
LEO~\cite{huang2023embodied} & 81.1 & 73.3 & 51.9\\
\midrule
MMCoT$^3$ & \textbf{83.9}& \textbf{84.1} & \textbf{54.3}\\
\bottomrule
\end{tabular}}
\end{table}

\renewcommand{\arraystretch}{1.1}
\vspace{-1mm}
\begin{table}[t]
\centering
\caption{Step-by-step ablation of MMCoT$^3$. We progressively build the full model from the instruction-tuned baseline by adding textual CoT supervision, 3D--Text interleaving, evidence fusion, and ambiguity modeling. ``EM@1" denotes top-1 exact match accuracy. Scan2Cap n-gram metrics use IoU $\ge$ 0.5. “Sim” denotes sentence similarity.}
\vspace{-2mm}
\label{tab:ablation_main}
\resizebox{\linewidth}{!}{
\begin{tabular}{lccc}
\toprule
Method & Scan2Cap & SQA3D & 3RScanPlanning \\
\cmidrule(lr){2-2} \cmidrule(lr){3-3} \cmidrule(lr){4-4}
 & CIDEr & EM@1 & Sim \\
\midrule
Instruction-tuned baseline & 69.2 & 52.7 & 53.9 \\
+ Textual CoT supervision & 73.4 & 59.6 & 69.7 \\
+ Naive 3D--Text interleaving & 75.7 & 55.3 & 66.1 \\
+ Evidence fusion & 76.9 & 58.4 & 78.6 \\
+ Ambiguity modeling (Full MMCoT$^3$) & \textbf{79.4} & \textbf{65.4} & \textbf{83.9} \\
\bottomrule
\end{tabular}}
\end{table}

\subsection{Visual Analysis}
We compare MMCoT$^3$ with an instruction-tuned baseline in Fig.~\ref{fig:view}. The baseline shares the same backbone but does not use textual CoT supervision, 3D--Text interleaving, evidence fusion, or ambiguity modeling. In Fig.~\ref{fig:view}(a) and (b), MMCoT$^3$ correctly rejects the queried \textit{camisole} and \textit{hard hat}, whereas the baseline hallucinates nonexistent objects. This suggests that explicit 3D--Text interleaving enables the model to verify intermediate conclusions against actual scene evidence, rather than relying on text-centric priors. In Fig.~\ref{fig:view}(c) and (d), the baseline further struggles with instance-level ambiguity. It fails to correctly identify the queried side table and is misled by a visually similar nearby bench, leading to an incorrect color prediction in (c), and it confuses the printer on the table with a similarly shaped microwave in (d). By contrast, MMCoT$^3$ resolves these ambiguities and produces the correct answers, which we attribute to the integration of task-relevant 3D evidence with ambiguity-aware evidence aggregation. These examples further demonstrate that MMCoT$^3$ improves both hallucination suppression and fine-grained instance disambiguation in complex 3D scenes.

\subsection{Ablation Study}

We conduct ablation studies to assess the contribution of each core component in MMCoT$^3$, including textual CoT supervision, 3D--Text interleaving, evidence fusion, and ambiguity modeling. As shown in Table~\ref{tab:ablation_main}, introducing textual CoT supervision already yields clear gains over the instruction-tuned baseline on all three benchmarks, improving Scan2Cap from 69.2 to 73.4 CIDEr, SQA3D from 52.7 to 59.6 EM@1, and 3RScanPlanning from 53.9 to 69.7 Sentence Similarity. These improvements confirm the importance of explicit reasoning supervision for 3D scene understanding.

Adding naive 3D--Text interleaving leads to mixed results: it further improves Scan2Cap from 73.4 to 75.7 CIDEr, but degrades SQA3D from 59.6 to 55.3 EM@1 and 3RScanPlanning from 69.7 to 66.1 Sim. This suggests that direct mention-to-object alignment is brittle in realistic 3D scenes, especially when multiple instances share similar semantics. Introducing evidence fusion substantially recovers the performance, particularly on 3RScanPlanning, where the score rises from 66.1 to 78.6, demonstrating the benefit of aggregating multiple candidate object tokens rather than prematurely committing to a single one. Finally, adding ambiguity modeling further improves all three benchmarks and yields the full MMCoT$^3$, achieving the best results of 79.4 CIDEr, 65.4 EM@1, and 83.9 Sim. These results show that both evidence fusion and ambiguity modeling are crucial for robust 3D--Text interleaved reasoning.

\section{Conclusion}

In this paper, we presented MMCoT$^3$, the first 3D--Text interleaved Chain-of-Thought framework for point-cloud-based 3D scene understanding. By explicitly interleaving intermediate textual reasoning with task-relevant 3D evidence, MMCoT$^3$ moves beyond the text-centric inference paradigm of prior 3D LLMs and enables active evidence consultation throughout multi-step reasoning. 
To further address the brittleness of direct object--mention alignment in realistic 3D scenes, we introduced an ambiguity-aware evidence aggregation mechanism that fuses candidate object features and injects both evidence and ambiguity cues into the reasoning process. 
We also developed PointCoT, a large-scale 3D-specific CoT dataset that provides explicit supervision for grounded multi-step reasoning across diverse 3D understanding settings. Extensive experiments on six benchmarks built on ScanNet and 3RScan demonstrate the effectiveness and generality of MMCoT$^3$ across captioning, question answering, embodied reasoning, dialogue, and planning. We hope this work will inspire future research on more reliable, evidence-grounded, and interpretable 3D multimodal reasoning.

\begin{acks}
This paper is supported by the National Natural Science Foundation of China (62422610, U23A20386, 62276045, 62576072, 62441231). It is also supported by the Liao Ning Science and Technology Plan (2025JH2/\allowbreak 101330121, 2025JH2/\allowbreak 101330124, 2023JH26/\allowbreak 10200016) and the Dalian City Science and Technology Innovation Fund No. 2023JJ11CG001.
\end{acks}

\bibliographystyle{ACM-Reference-Format}
\bibliography{sample-base}

\end{document}